\documentclass[lettersize,journal]{IEEEtran}
\usepackage{amsmath,amsfonts}
\usepackage{algorithmic}
\usepackage{algorithm}
\usepackage{array}
\usepackage[caption=false,font=normalsize,labelfont=sf,textfont=sf]{subfig}
\usepackage{textcomp}
\usepackage{stfloats}
\usepackage{url}
\usepackage{verbatim}
\usepackage{graphicx}
\usepackage{cite}
\usepackage{multicol, multirow}
\usepackage{booktabs}
\usepackage{colortbl}
\usepackage[dvipsnames]{xcolor}
\usepackage{url}
\usepackage{color}
\definecolor{green1}{RGB}{84,130,53}
\definecolor{yellow1}{RGB}{255,192,0}
\definecolor{blue}{RGB}{133,151,176}
\definecolor{green2}{RGB}{84,130,53}
\definecolor{red}{RGB}{192,0,0}
\definecolor{orange}{RGB}{237,125,49}
\definecolor{mygray}{RGB}{220,220,220}

\usepackage{hyperref}
\hypersetup{
    colorlinks=true,
    linkcolor=NavyBlue,
    filecolor=black,      
    urlcolor=black,
    citecolor=NavyBlue,
    }

\begin{document}

\title{Knowledge-aware Text-Image Retrieval for Remote Sensing Images}

\author{Li Mi,~\IEEEmembership{Student Member,~IEEE,}
        Xianjie Dai,
        Javiera Castillo-Navarro,
        Devis Tuia,~\IEEEmembership{Fellow,~IEEE}
\thanks{This work was supported by EPFL Science Seed Funds under Grant 21692. L. Mi, X. Dai, J. Castillo-Navarro and D. Tuia are with École Polytechnique Fédérale de Lausanne (EPFL). Corresponding author: Li Mi (li.mi@epfl.ch).}}

\markboth{Journal of \LaTeX\ Class Files,~Vol.~14, No.~8, August~2021}%
{Shell \MakeLowercase{\textit{et al.}}: A Sample Article Using IEEEtran.cls for IEEE Journals}


\maketitle

\begin{abstract}
Image-based retrieval in large Earth observation archives is challenging because one needs to navigate across thousands of candidate matches only with the query image as a guide. By using text as information supporting the visual query, the retrieval system gains in usability, but at the same time faces difficulties due to the diversity of visual signals that cannot be summarized by a short caption only. For this reason, as a matching-based task, cross-modal text-image retrieval often suffers from information asymmetry between texts and images. To address this challenge, we propose a Knowledge-aware Text-Image Retrieval (KTIR) method for remote sensing images. By mining relevant information from an external knowledge graph, KTIR enriches the text scope available in the search query and alleviates the information gaps between texts and images for better matching. Moreover, by integrating domain-specific knowledge, KTIR also enhances the adaptation of pre-trained vision-language models to remote sensing applications. Experimental results on three commonly used remote sensing text-image retrieval benchmarks show that the proposed knowledge-aware method leads to varied and consistent retrievals, outperforming state-of-the-art retrieval methods.
\end{abstract}



\section{Introduction}

Recent advances in satellite data acquisition and storage have led to a rapid increase in the size and complexity of remote sensing image archives. To explore these archives, image retrieval has received increasing attention with multiple systems designed to conduct searches based on visual similarity to the query~\cite{zhou2018patternnet, hoxha2020toward}. However, retrieving images using example images limits the versatility of the retrieval system, since with the query image only, one cannot specify which elements are essential or what the retrieval objective is. As a solution, text-image retrieval~\cite{faghri2017vse, yuan2022remote, al2022multilanguage} has been introduced to explicit the retrieval targets in a semantic way. Text-image retrieval aims at finding an image based on a text or, in reverse, retrieving a text pertaining to an image.

\begin{figure}
    \centering
    \includegraphics[width=0.9\linewidth]{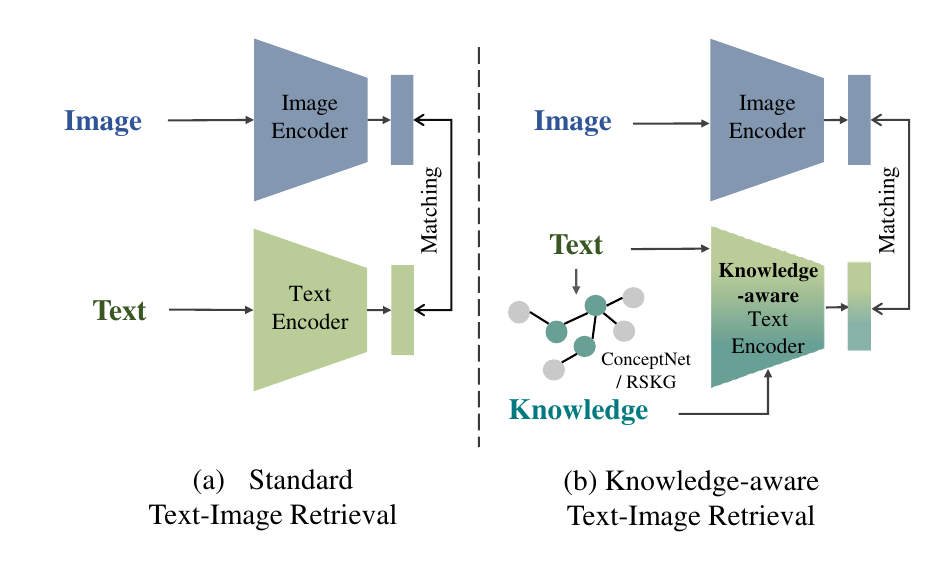}
    \caption{Intuition behind the proposed KTIR system: in a standard text-image retrieval approach (a), text and images are matched directly, while in KTIR (b), commonsense knowledge is added from external sources (a knowledge base) to make the retrieval more varied, robust to ambiguities and consistent with general knowledge.}
    \label{fig:motivation}
\end{figure}

When using text, the prospective retrieval system gains in usability, but at the same time faces the problem of information gaps between texts and images~\cite{wang2016comprehensive}. Previous efforts attempted to fill the cross-modal information gap by establishing a representative text-image joint embedding space. For example, recent works have explored image feature representations~\cite{yuan2021exploring, yuan2022remote, al2022multilanguage, yao2022hypergraph}, fusion models~\cite{lv2021fclm, mikriukov2022deep} and contrastive objectives~\cite{cheng2021deep, zheng2022scale} for better aligning the cross-modal features.

However, the information asymmetry caused by cross-modality information is rarely mentioned, \textit{i.e.}, the fact that the short texts do not allow much freedom to represent diverse image content. When dealing with very high-resolution remote sensing images, the image content can be very diverse, hence it is difficult to be comprehensively summarized by a short caption only. On one hand, human captions can only describe one or a few aspects of the image, focusing on the most dominant information. For example, one image could receive the following caption text: \textit{There is a lake}. Nevertheless, there might be trees and mountains around the lake which are ignored by humans or caption generators. On the other hand, different people will describe the image from subjective perspectives, resulting in a variety of text information for a single image, which may confuse the matching model. For example, the captions \textit{There is a lake} and \textit{There are many boats} could refer to the same image. Therefore, strategies to handle lacunary captions, nuances and synonyms are needed for the task, and a balance between objectivity and completeness must be achieved. 

Commonsense knowledge sources have been recognized as effective priors in many vision-and-language research~\cite{yang2018visual, fang2017object} to reveal commonsense and alleviate ambiguities. For example, by adding commonsense knowledge from external sources, the description \textit{This is a lake} will probably be expanded by introducing related concepts and relations such as \textit{Lake has water} and \textit{Boats on the lake}, which expand the text content to match the image. In addition, external knowledge also provides an opportunity to link descriptions of different concepts. For example, the concepts \textit{lake} and \textit{boat} could be bridged by \textit{Boat on the lake} from external knowledge sources.

In order to fill the cross-modal information asymmetry and reduce the impact of language ambiguity, we propose a Knowledge-aware Text-Image Retrieval (KTIR) method to introduce external knowledge for remote sensing images into text-image retrieval tasks (Fig.~\ref{fig:motivation}). 
More specifically, based on the objects mentioned in a sentence as starting points, KTIR proposes to mine the expanded nodes and edges in the external knowledge graph and embed them as features to enrich those extracted from the text content alone. In a way, KTIR adds extra commonsense-based links which can enrich the semantics, expand the scope of the text query, alleviate the potential language ambiguity and also facilitate the adaptation of the general vision-language pretraining models to the remote sensing domain. To demonstrate the effectiveness of the proposed KTIR method, we design experiments on three commonly used remote sensing text-image retrieval benchmarks: the UCM-Caption dataset~\cite{qu2016deep}, the RSICD dataset~\cite{lu2017exploring}, and the RSITMD dataset~\cite{yuan2021exploring}. Results show that KTIR outperforms the comparison methods. Our experiments also explore the differences between different knowledge sources and the relationship between knowledge and image content.

The remainder of the paper is organised as follows: Section~\ref{sec:related_work} details the related works about remote sensing text-image retrieval, external knowledge sources and knowledge-aware vision-language research. Section~\ref{sec:method} presents the proposed KTIR method. Section~\ref{sec:setup} and Section~\ref{sec:result} present the experimental settings and the experiment results, respectively. Section~\ref{sec:conclusion} concludes the paper.


\section{Related Work}
\label{sec:related_work}
\subsection{Text-Image Retrieval for Remote Sensing Images}
Due to the increasing quantity of multi-modal remote sensing data, vision-language research, such as image captioning~\cite{lu2017exploring}, visual question answering~\cite{lobry2020rsvqa}, text-guided visual grounding and cross-modal retrieval~\cite{yuan2022remote} has attracted increasing attention~\cite{tuia2021toward} in remote sensing. Among them, text-image retrieval is regarded as one of the fundamental vision-language tasks for cross-modal alignment. Recent advances in remote sensing text-image retrieval mainly focused on building a representative joint embedding space, especially a multi-level or multi-scale image representation~\cite{yuan2021exploring, yuan2022remote, yuan2021lightweight, pan2023swan}. For example, Yuan \textit{et al.}~\cite{yuan2022remote} designed a dynamic fusion module of the global and local information to generate a multi-level visual representation. Yuan \textit{et al.}~\cite{yuan2021exploring} designed an asymmetric multimodal feature matching network which can adapt to multi-scale image inputs. Some works mentioned the information asymmetry between image and text and tried to address it by extracting more representative features~\cite{yu2022text, yao2022hypergraph, al2022multilanguage} or designing a powerful fusion method~\cite{lv2021fclm, cheng2021deep, yuan2022mcrn, tang2023interacting}. For example, Yu \textit{et al.}~\cite{yu2022text} proposed to use a Graph Neural Network (GNN) for better representing the object relationships in the image content. Cheng \textit{et al.}~\cite{cheng2021deep} proposed an attention-based module to fuse the feature from different modalities and used a triplet loss to learn the matching.

Besides the aforementioned methods that are proposed especially for text-image retrieval, vision-language models in remote sensing~\cite{liu2023remoteclip, zhang2023rs5m, silva2024large, yuan2023parameter} also regard text-image retrieval as a fundamental training and evaluation task. To train those models, a large amount of domain-specific annotations is crucial to the adaptation of pre-trained models to remote sensing images. For example, RemoteCLIP~\cite{liu2023remoteclip} is trained on 17 remote sensing datasets including detection, segmentation and text-image retrieval. GeoRSCLIP~\cite{zhang2023rs5m} collected an additional dataset of 5 million remote sensing images for pretraining. 

Despite the tremendous progress made, the insufficiency and ambiguity of textual information are rarely addressed. Departing from previous efforts, which based the retrieval on the image and caption only, we propose to enrich the latter with external knowledge sources that would extend the text content and alleviate ambiguities. Moreover, external knowledge~\cite{speer2017conceptnet, li2021robust} also serves as a bridge to adapt pre-trained general vision-language models~\cite{li2022blip, radford2021learning} to the remote sensing domain without supplementary annotations. 

\subsection{External Knowledge Sources}
Sources of external knowledge can be wide and diverse, including different types of knowledge. In vision-language research, commonsense knowledge~\cite{speer2017conceptnet} and domain-specific knowledge \cite{li2021robust} are often used. 
Commonsense knowledge bases (\textit{e.g.}, ConceptNet~\cite{speer2017conceptnet} and ATOMIC~\cite{sap2019atomic}) include the basic concepts and facts which are usually shared by most people and implicitly assumed in communications. For example, everyday events and their effects (\textit{e.g.}, \textit{eat something if feeling hungry}), facts about beliefs and desires (\textit{e.g.}, \textit{keep exercising to get in good health}), and properties of objects (\textit{e.g.}, \textit{fishes live in water}). Most of the commonsense knowledge sources use triplets (\textit{i.e.}, $<$head, relation, tail$>$) to store and represent knowledge. For example, the triplet $<$\textit{cooking, Requires, food}$>$ means that `\textit{the prerequisite of cooking is food}'. In this paper, we consider ConceptNet~\cite{speer2017conceptnet} as a commonsense knowledge source. ConceptNet is a multilingual knowledge graph that aligns its knowledge resources to 36 types of relations, including symmetric relations (\textit{e.g.}, \textit{RelatedTo} and \textit{SimilarTo}) and asymmetric relations (\textit{e.g.}, \textit{AtLocation}, \textit{CapableOf}, \textit{Causes}, \textit{Desires}, \textit{HasA}, \textit{HasProperty}, \textit{UsedFor}, \textit{etc}.).

Different from commonsense knowledge that is generic to many domains and daily life, domain-specific knowledge is adapted to a particular domain. The domain-specific knowledge is obtained by filtering out irrelevant objects and relations from general knowledge bases or directly collecting from domain-specific corpus or annotations~\cite{li2021robust, sun2022remote}. In remote sensing research, Li \textit{et al.}~\cite{li2021robust} constructed a remote sensing knowledge graph (RSKG) to support zero-shot remote sensing image scene classification. RSKG has 117 entities, 26 relations and 191 triples, manually selected for remote sensing images. We consider RSKG as the domain-specific knowledge source in the paper. Detailed information on the knowledge sources we used in the paper can be found in Section \ref{sec:knowledge_info}.

\begin{figure*}[t]
    \centering
    \includegraphics[width=0.95\linewidth]{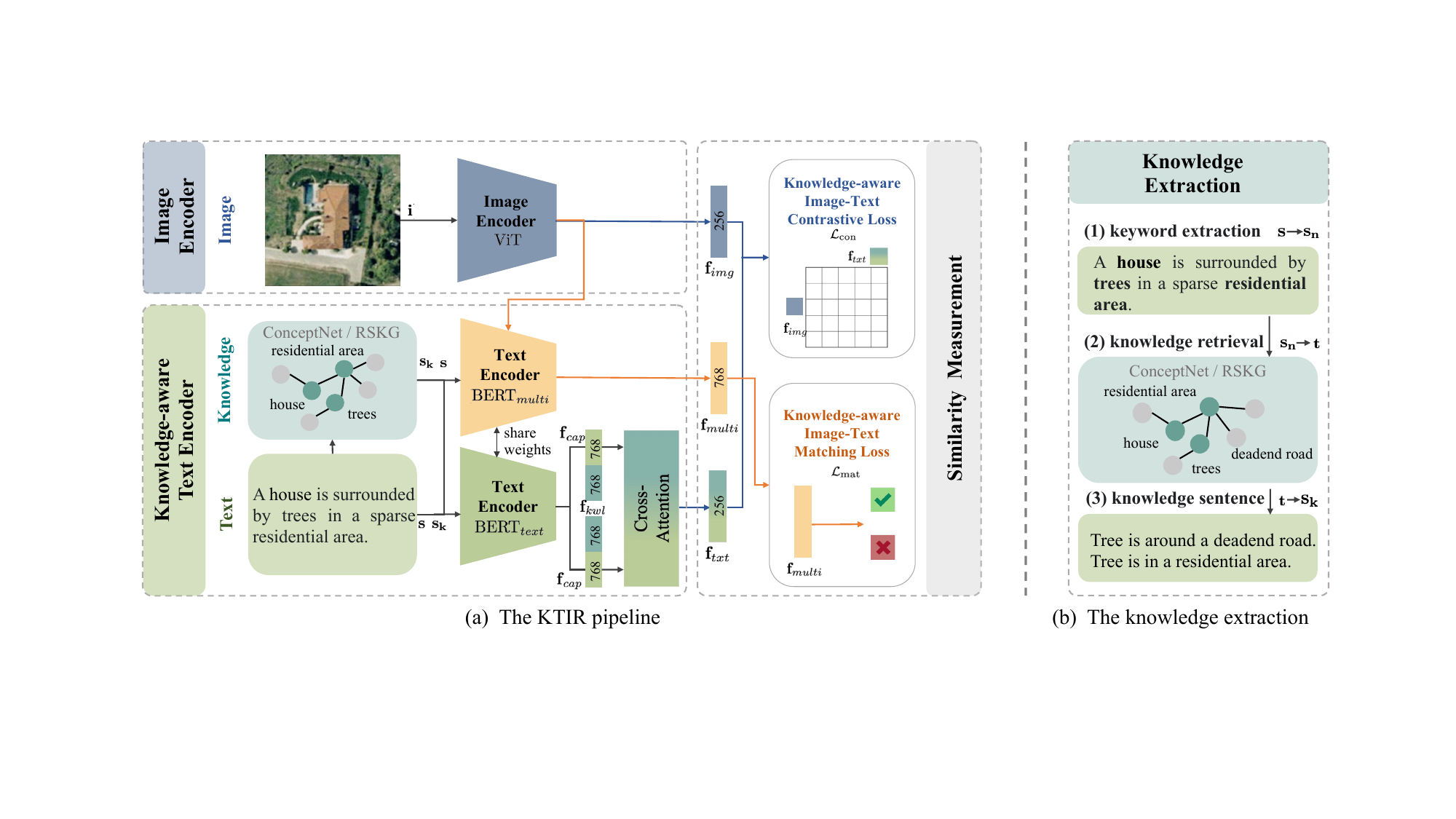}
    \caption{(a) The pipeline of the KTIR. The proposed text-image retrieval system comprises three main components: an image encoder, a knowledge-aware text encoder and a similarity measurement module. The image feature ($\mathbf{f}_{img}$) are obtained by ViT~\cite{dosovitskiy2020image} (\textcolor{blue}{$\operatorname{ViT}$}). A BERT~\cite{devlin2019bert} is used as text-only mode (\textcolor{green1}{$\operatorname{BERT}_{text}$}, green mode) and multimodal mode (\textcolor{yellow1}{$\operatorname{BERT}_{multi}$}, yellow mode) to encode the knowledge-aware text feature ($\mathbf{f}_{txt}$) and the multimodal feature ($\mathbf{f}_{multi}$), respectively. Then the text-image contrastive loss ($\mathcal{L}_{\mathrm{con}}$) and the text-image matching loss ($\mathcal{L}_{\mathrm{mat}}$) are used as the training objectives for cross-modality retrieval. (b) The knowledge extraction process. In the knowledge-aware text encoder, the knowledge extraction process includes keyword extraction, knowledge retrieval and knowledge sentence construction. After knowledge extraction, the knowledge sentences $\mathbf{s_k}$ are collected for each caption $\mathbf{s}$. Numbers in the feature vectors denote their dimension.}
    \label{fig:framework}
\end{figure*}

\subsection{Knowledge-aware Vision-Language Research}

Explicitly incorporating knowledge into language models has been an emerging trend in recent natural language processing (NLP) research~\cite{hu2023survey, yin2022survey}. Similar to the pure NLP tasks, recent research has shown that many vision tasks such as visual question answering, image captioning, and vision-language navigation, can be enhanced by adding knowledge~\cite{wang2017fvqa, yang2018visual, wei2021knowledge, gui2022kat, hu2023reveal, shi2019knowledge, wang2020consensus}. In those tasks, external knowledge is often regarded as a significant source of information that is difficult to obtain directly from vision. For example, to answer knowledge-aware visual-related questions~\cite{wang2017fvqa}, the model is supposed to understand the vision content, as well as to retrieve the knowledge bases to obtain concepts and relations that are not visible in the image itself.

In this work, external knowledge is integrated into text-image retrieval in remote sensing images to 1) narrow the cross-modality information gap by the explicit integration of external knowledge; and 2) adapt general vision-language models to the remote sensing domain by using domain-specific commonsense knowledge. 
The proposed KTIR is an extension of our preliminary work KCR~\cite{mi2022knowledge}. The differences between KCR and KTIR are as follows: 
\begin{itemize}
    \item \textit{The knowledge sources}: KCR only supports using RSKG as knowledge sources while KTIR also includes ConceptNet (See Section~\ref{sec:knowledge}).
    \item \textit{The base models}: The text encoder in KCR is frozen while KTIR is based on the BLIP model~\cite{li2022blip} and all the modules are trainable.
    \item \textit{Knowledge embedding methods}: KCR combines the knowledge triplets and the captions directly, while KTIR uses a cross-attention mechanism to fuse two text sources (See Section~\ref{sec:ablation}). 
\end{itemize}


\section{Knowledge-aware Text-Image Retrieval Method}
\label{sec:method}
The proposed text-image retrieval system comprises three main components: an image encoder, a knowledge-aware text encoder and a similarity measurement module (Fig.~\ref{fig:framework} (a)). The image encoder is designed to extract image features by a Vision Transformer (ViT)~\cite{dosovitskiy2020image}. The text encoder embeds a sentence and its related external knowledge into a joint feature space representing the text inputs. Finally, the image and text features are both used within the similarity measurement module to compute the similarity score between text queries and candidate images, which are then ranked according to their relevance. The model can also be applied in reverse, where the best captions to summarize an image are retrieved. Note that we build the KTIR method upon the BLIP~\cite{li2022blip} model, but different backbones could be used instead (See Section~\ref{sec:backbone}).

\subsection{Image encoder}

\noindent The image encoder is a ViT, where an image $\mathbf{i}$ is divided into several patches, which are processed by a transformer encoder. Then, the extracted image features are further reprojected to a space of dimension equivalent to the output of the text encoder described in the next section. A fully-connected (FC) layer is used for such reprojection. The image embedding process can be denoted as:
\begin{equation}
    \mathbf{f}_{img} = \operatorname{FC}_{img} \mathit{(\operatorname{ViT}(\mathbf{i}))}.
\end{equation}

\subsection{Knowledge-aware text encoder}
\label{sec:knowledge_info}
\noindent The knowledge-aware text encoder embeds the textual description for text-image retrieval. It uses as input a number of captions describing the scene as well as external knowledge to strengthen the text representation. In this section, we first explain the knowledge extraction process (Fig.~\ref{fig:framework} (b)) and then describe the fusion of captions and external knowledge in the text encoder.

\vspace{0.7em}
\subsubsection{\textbf{Knowledge extraction}}

The key idea is to retrieve and represent relevant knowledge triplets from external knowledge sources based on the information provided by the caption sentence. The knowledge extraction is repeated for each caption and consists of the following three steps:
\paragraph{Keyword extraction}
For a caption $\mathbf{s}$ with $n$ words: $\mathbf{s}=\{w_1, w_2, ..., w_n\}$ $(n \geq 1)$, a tokenizer is used to separate every word and divide the part-of-speech (\textit{e.g.}, noun, verb, adjective, adverb, \textit{etc}.). Based on the part-of-speech tags, all the nouns can be appended to a word list $\mathbf{s_n}$. Note that we turn all the plurals into their singular form.

\paragraph{Knolwedge triplet retrieval}
We use two different knowledge sources, RSKG~\cite{li2021robust} and ConceptNet~\cite{speer2017conceptnet}. The combination of the two sources is also considered.
\begin{itemize}
    \item {RSKG}~\cite{li2021robust} is a hand-crafted knowledge graph for the remote sensing domain. The knowledge graph is designed for the remote sensing scenes so that the objects and relations conform to remote sensing vocabulary. Based on the keyword list $\mathbf{s_n}$, we retrieve all the related knowledge triplets in the graph $\mathbf{t_{rs}}=\{t_{rs}^{1}, t_{rs}^{2}, ..., t_{rs}^{m_{rs}}\}$, where $m_{rs}$ is the number of triplets from the RSKG knowledge graph and $t_{rs}^{m_{rs}}$ is a triplet $<$head, relation, tail$>$ and the head or the tail is in the keyword list $\mathbf{s_n}$. More specifically, the nouns in the word list are regarded as the initial nodes. Starting from those nodes, all the one-step neighbours with the connected edges in RSKG are included. We keep all the relations from the graph.
    \item ConceptNet~\cite{speer2017conceptnet} is a multi-language commonsense knowledge graph. Compared to RSKG, ConceptNet is much larger and more general in terms of objects and relationships. Given a keyword, the ConceptNet official API~\footnote{https://conceptnet.io/} returns the related triplets that involve the query keywords. Similar to RSKG, a triplet list $\mathbf{t_{ce}}=\{t_{ce}^{1}, t_{ce}^{2}, ..., t_{ce}^{m_{ce}}\}$ is constructed based on the word list $\mathbf{s_n}$. $m_{ce}$ is the number of triplets from ConceptNet. Note that during the retrieval, we filter out the triplets that have non-English words. For the relations, we choose 15 relations from the total set, they are: \textit{UsedFor, ReceivesAction, HasA, Causes, HasProperty, CreatedBy, DefinedAs, AtLocation, HasSubEvent, MadeUpOf, HasPrerequisite, Desires, NotDesires, IsA} and \textit{CapableOf}.
    \item Combine RSKG and ConceptNet. For the triplet list obtained from ConceptNet $\mathbf{t_{ce}}$, we filter out items where both head and tail are not in the RSKG objects to limit the external knowledge to remote sensing domain. Together with the RSKG triplet list $\mathbf{t_{rs}}$, a new triplet list $\mathbf{t_{co}}=\{t_{co}^{1}, t_{co}^{2}, ..., t_{co}^{m_{co}}\}$, is constructed by combining the triplets from two knowledge sources, where $m_{co} \leq m_{re} + m_{ce}$.
\end{itemize}
Detailed statistics of different external knowledge sources are shown in Table~\ref{tab:sources}. In general, $\mathbf{t_{rs}}$ has fewer objects and relation types compared to $\mathbf{t_{ce}}$ and $\mathbf{t_{co}}$. By combining two knowledge sources ($\mathbf{t_{co}}$), a reasonable amount of additional concepts and diverse types of relations are considered. Note that mining all the related objects and relations might be redundant, so in the experiments, we randomly select $m$ triplets from all the available ones for a caption $\mathbf{s}$ (Detailed analysis of triplet numbers and selection strategies can be found in Section~\ref{sec:ablation} and Section~\ref{ssec:triplet_selection}, respectively). In the end, the selected triplet list can be denoted as $\mathbf{t}$.

\paragraph{Knowledge sentence construction}
Before encoding the knowledge triplets, we convert the selected ones (composing the list $\mathbf{t}$) to short knowledge sentences. More specifically, for a triplet ($<$head, relation, tail$>$), we keep the head subject and tail object as they are, but re-formulate the relation to construct a meaningful sentence according to a transformation template (\textit{e.g.}, \textit{UsedFor} as ``\textit{is used for}''). The specific templates are listed in Table~\ref{tab:template}. Following these rules, for example, the triplet $<$ boat, AtLocation, water $>$ can be re-written as \textit{boat is at location of water}. After this re-formulation, a list of knowledge sentences is obtained, one per knowledge triplet in list $\mathbf{t}$. We then combine $m$ knowledge sentences as the final knowledge sentence for caption $\mathbf{s}$, which we refer to as $\mathbf{s_k}$.

\begin{table}[t]
\renewcommand{\arraystretch}{1.2}
\centering
\caption{The statistics of the number of objects (\# Objects), relations (\# Relations) and triplets (\# Triplets) in different knowledge sources.}
\resizebox{0.88 \columnwidth}{!}{
\begin{tabular}{ccccc}
\hline
Knowledge Sources & \# Objects & \# Relations & \# Triplets\\
\hline 
RSKG~\cite{li2021robust} ($\mathbf{t_{rs}}$) & 117 & 26 & 191 \\
ConceptNet~\cite{speer2017conceptnet} ($\mathbf{t_{ce}}$) & 3855 & 15 & 3343 \\
Combined ($\mathbf{t_{co}}$) & 908 & 41 & 748 \\
\hline
\end{tabular}
}
\label{tab:sources}
\end{table}

\begin{table}[t]
\caption{Templates used to convert relations in the knowledge triplets into knowledge sentences.}
\label{tab:template}
\setlength{\tabcolsep}{3pt}
\centering
\begin{tabular}{llll}
\hline
{Relation} & {Template}  & {Relation} & {Template}  \\
\hline
UsedFor         & is used for         &  shape & shape is \\
ReceivesAction  & receives action     &  color & color is \\
HasA            & has a               &  width & width is \\
Causes          & causes              & distribution & distribution is \\
HasProperty     & has a property      & height & height is \\
CreatedBy       & is created by       & next\_to & next to\\
DefinedAs       & is defined as       & stop\_at & stop at\\
AtLocation      & is at location of   & pass\_through & pass through\\
HasSubEvent     & has                 & intersect\_at & intersect at \\
MadeUpOf        & is made of          & marked\_on & is marked on\\
HasPrerequisite & has prerequisite to & connected\_to & is connected to\\
Desires         & desires             & is\_component\_of & is component of\\
NotDesires      & not desires         & is\_part\_of & is part of\\
IsA             & is a                & is\_member\_of & is member of\\
CapableOf       & is capable of       &  & \\
\hline
\end{tabular}

\end{table}

\subsubsection{\textbf{Fusion in the text encoder}} Same as BLIP~\cite{li2022blip}, a BERT~\cite{devlin2019bert}-based structure is used as text encoder. According to the BLIP setting, the encoder can be used in two different modes: the text-only mode and the multimodal mode. Here, we use the text-only mode ($\operatorname{BERT}_{text}$) to represent textual features including the caption embedding and the knowledge sentence embedding (See Eq.~\ref{eq:text}). This text-only embedding is used to calculate text-image contrastive loss. The multimodal mode ($\operatorname{BERT}_{multi}$) will be used to fuse the image features with the text features (See Eq.~\ref{eq:multif}) and will be used for a multimodal matching loss in the following sections.

\paragraph{knowledge-aware text feature} The caption sentences and the knowledge sentences are passed into the text encoder to get the caption representation ($\mathbf{f}_{cap}$) and the knowledge representation ($\mathbf{f}_{kwl}$), specifically.
\begin{equation}
\label{eq:text}
    \begin{aligned}
     \mathbf{f}_{cap} & = \operatorname{BERT}_{text} \mathit{(\mathbf{s})}, \\
     \mathbf{f}_{kwl} & = \operatorname{BERT}_{text} \mathit{(\mathbf{s_k})}.\\
    \end{aligned}
\end{equation}
The knowledge-aware text feature is generated by fusing $\mathbf{f}_{cap}$ and $\mathbf{f}_{kwl}$ through a single cross-attention layer. The cross-attention mechanism can be represented as:
\begin{equation}
\operatorname{CrossAtt}(\mathbf{f}_{1}, \mathbf{f}_{2})=\operatorname{softmax}\left(\frac{\mathbf{f}_{1}W_{1} (\mathbf{f}_{2}W_{2})^T}{\sqrt{d_k}}\right) \mathbf{f}_{2}W_{2},
\end{equation}
where $d_k$ is the dimension of the feature space, and $W_{1} \in \mathbb{R}^{d_k \times d_k}$ and $W_{2} \in \mathbb{R}^{d_k \times d_k}$ are the transformation matrices for input features $\mathbf{f}_{1} \in \mathbb{R}^{d_k}$ and $\mathbf{f}_{2} \in \mathbb{R}^{d_k}$, respectively. We input both sequences (caption representations, then knowledge sentence representations and their reverse) as the input of the cross-attention layer. Note that after concatenation, the dimension of the concatenated sequences is reduced to $d_k$. After being embedded in the feature space, a final FC layer is applied to obtain the overall representation of a text:
\begin{equation}
\label{eq:textf}
     \mathbf{f}_{txt} = \operatorname{FC}_{text} ( \mathit{\operatorname{CrossAtt}([\mathbf{f}_{cap}, \mathbf{f}_{kwl}],[\mathbf{f}_{kwl},\mathbf{f}_{cap}])}),
\end{equation}
where [ , ] denotes the concatenation operation and the FC layer to reduce the feature dimensions.
\paragraph{multi-modal feature} As stated in the previous subsection, the BLIP text encoder can also be used in multimodal mode ($\operatorname{BERT}_{multi}$), where the text inputs ($\mathbf{s,s_k}$) and image embeddings before $\operatorname{FC}_{img}$ layer, denoted as $\mathbf{f}_{img}'$, are used as the two inputs of the text encoder. The multimodal mode builds a joint feature ($\mathbf{f}_{multi}$) for multimodal inputs: 
\begin{equation}
\label{eq:multif}
    \mathbf{f}_{multi} = \operatorname{BERT}_{multi} \mathit{(\mathbf{(s,s_k)}, \mathbf{f}_{img}')}.
\end{equation}
The multimodal feature is then used in a FC layer ($\operatorname{FC}_{multi}$) acting as a classifier to compute the probability of alignment between the image and text pair. This is implemented as a binary classification task:
\begin{equation}
\label{eq:equation}
    \hat{y}=\operatorname{FC}_{multi} (\mathbf{f}_{multi}), 
\end{equation}
where $\hat{y}$ denotes the prediction for the text-image pair, indicating whether they are matched or not.

\subsection{Similarity Measurement}
After encoding, the text and image are represented as image feature ($\mathbf{f}_{img}$), knowledge-aware text feature ($\mathbf{f}_{txt}$), multi-modal joint embedding ($\mathbf{f}_{multi}$) and a predicted probability of matching ($\hat{y}$). In this section, we describe the objectives based on those features to perform cross-modal retrieval.

\vspace{0.7em}
\subsubsection{\textbf{Knowledge-aware text-image contrastive loss}} The contrastive loss constrains the similarity score of the matched image-text pairs to be higher than the similarity score of the unmatched ones. It creates a joint feature embedding space for both image and text by aligning the feature embeddings of the paired images and texts. We construct the two contrastive losses for text-image matching ($\mathcal{L}_{\mathrm{img2txt}}$) and text-image matching ($\mathcal{L}_{\mathrm{txt2img}}$), respectively:

\begin{equation}
   \begin{aligned}
    \mathcal{L}_{\mathrm{img2txt}}=-\log \frac{\exp \left(\mathbf{f}_{img} \cdot \mathbf{f}_{txt}^{+} / \tau\right)}{\sum_{i=1}^N \exp \left(\mathbf{f}_{img} \cdot \mathbf{f}_{txt}^i / \tau\right)}, \\
    \mathcal{L}_{\mathrm{txt2img}}=-\log \frac{\exp \left(\mathbf{f}_{txt} \cdot \mathbf{f}_{img}^{+} / \tau\right)}{\sum_{i=1}^N \exp \left(\mathbf{f}_{txt} \cdot \mathbf{f}_{img}^i / \tau\right)}, \\
   \end{aligned}
\end{equation}
where $\mathbf{f}_{txt}^{+}$ and $\mathbf{f}_{img}^{+}$ represent the positive examples and $N$ is the number of pairs in a batch. $\tau$ is a temperature parameter. Finally, the contrastive loss is:
\begin{equation}
    \mathcal{L}_{\mathrm{con}}=\frac{1}{2} (\mathcal{L}_{\mathrm{img2txt}}+\mathcal{L}_{\mathrm{txt2img}}).
\end{equation}

\vspace{0.7em}
\subsubsection{\textbf{Knowledge-aware text-image matching loss}} Unlike the contrastive loss that aims to align the unimodal features, the matching loss learns fine-grained multimodal alignment by a binary classification task~\cite{li2021align, li2022blip}. The model uses a linear layer to predict whether an image-text pair is positive (matched) or negative (unmatched) given its multimodal feature.
The matching loss is the binary cross entropy loss:
\begin{equation}
    \mathcal{L}_{\mathrm{mat}}=-\frac{1}{N} \sum_{i=1}^N y_i \cdot \log \left(\hat{y_i}\right)+\left(1-y_i\right) \cdot \log \left(1-\hat{y_i}\right),
\end{equation}
where $y_i$ is the binary label depicting whther the $i$-th image-text pair is a match and $\hat{y_i}$ is the corresponding predicted probability from the binary classifier (Eq.~\ref{eq:equation}).

The final training objective $\mathcal{L}$ is the combination of knowledge-aware text-image contrastive loss and knowledge-aware text-image matching loss with the weights $w_1$ and $w_2$:
\begin{equation}
\label{eq:loss}
\mathcal{L}=w_1\mathcal{L}_{\mathrm{con}}+w_2\mathcal{L}_{\mathrm{mat}}.
\end{equation}
In practice, we follow the loss calculation described in previous work~\cite{li2021align, li2022blip}: A momentum encoder is introduced to create soft labels as training targets. This is to account for the potential positives in the negative pairs for knowledge-aware text-image contrastive loss ($\mathcal{L}_{\mathrm{con}}$). We use a hard negative sampling strategy~\cite{li2021align} to find negative samples showing the highest contrastive similarity in the mini-batch. We then use those samples for computing the knowledge-aware text-image matching loss ($\mathcal{L}_{\mathrm{mat}}$).

\subsection{Inference}
At inference, text and image features are obtained by the encoders. Two scores are used to decide the final retrieval result. Firstly, the text-image similarity score ($\mathrm{S}_{\mathrm{sim}}$) is calculated as the pairwise cosine similarity scores between image features and text features. In this case, the text encoder uses the text-only mode ($\operatorname{BERT}_{text}$). Secondly, a text-image matching score ($\mathrm{S}_{\mathrm{mat}}$) is obtained by the output probability of the binary classifier using the multi-modal mode of the text encoder ($\operatorname{BERT}_{multi}$). The final score $\mathrm{S}$ is calculated as a sum of the two scores:
\begin{equation}
\label{eq:score}
    \mathrm{S}=\mathrm{S}_{\mathrm{sim}} + \mathrm{S}_{\mathrm{mat}}.
\end{equation}
For text-image retrieval, both the text and image features are used to compute the final score between a text query and candidate images, which are then ranked according to their relevance. When applied in reverse, where the best captions to summarize an image are retrieved based on the similarity score between the query image and the candidate texts.


\section{Experimental Setup}
\label{sec:setup}
\subsection{Datasets}
\begin{figure}[t]
    \centering
    \includegraphics[width=0.95\linewidth]{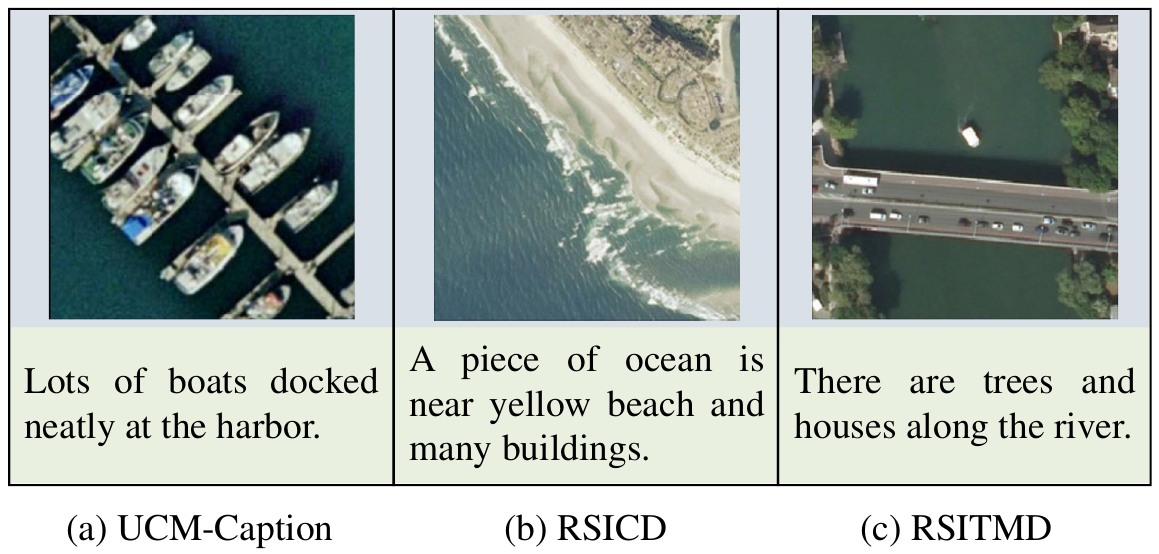}
    \caption{Examples of images and text sentences are from the three datasets.}
    \label{fig:dataset}
\end{figure}

We perform experiments on three RS text-image datasets: UCM-Caption~\cite{qu2016deep}, RSICD~\cite{lu2017exploring}, and RSITMD~\cite{yuan2021exploring}. Examples from the three datasets are shown in Fig~\ref{fig:dataset}.

\begin{itemize}

\item \textbf{UCM-Captions}~\cite{qu2016deep} is based on the UC Merced Land Use dataset~\cite{yang2010bag}. It contains remote sensing images categorized into 21 scene categories (\textit{e.g.}, buildings, intersection, parking lost, runway, agricultural, forest, \textit{etc}), with 100 samples for each class. Each image contains 256 $\times$ 256 pixels and 5 sentences annotated.

\item \textbf{RSICD}~\cite{lu2017exploring} is a large remote sensing text-image dataset, and a commonly used benchmark for remote sensing text-image retrieval and remote sensing image captioning. It contains 10921 images with the size 224$\times$224 pixels with various resolutions. There are also 5 sentences per image.

\item \textbf{RSITMD}~\cite{yuan2021exploring} is a fine-grained remote sensing text-image dataset. It contains 4743 images from 24 categories with 23715 captions and 21403 keywords. Compared to the RSICD dataset, the RSITMD dataset was designed to have more fine-grained and diverse text descriptions.

\end{itemize}

We follow the train-test split in previous work \cite{yuan2022remote, yuan2021exploring, zhang2023rs5m}.

\subsection{Metrics} 
To evaluate the model performance, we exploit the standard evaluation metrics in retrieval tasks and measure the rank-based performance by R@$k$ and mR~\cite{faghri2017vse, huang2018deep, yuan2021exploring}. With different values of $k$, R@$k$ means the fraction of queries for which the most relevant item is ranked among the top-$k$ retrievals. mR represents the average of all R@$k$ in both text-image retrieval and image-text retrieval. mR$_{t2i}$ and mR$_{i2t}$ denote the average of all R@$k$ in text-image retrieval and image-text retrieval, respectively. In our experiments, we report the results of $k = [1, 5, 10]$, as in previous works.

\subsection{Implementation details} 
Following BLIP, the image encoder is a ViT-B/16, a ViT architecture with 12 attention heads, $12$ hidden layers, and images divided into $16\times16$ patches. The text encoder is $\text{BERT}_{\text{base}}$, a transformer encoder with $12$ attention heads and 12 hidden layers. The output feature dimension of both ViT and $\text{BERT}_{\text{base}}$ is $768$. The dimension of the final image feature ($\mathbf{f}_{img}$), of the knowledge-aware text feature ($\mathbf{f}_{txt}$) and of the multi-modal joint embedding ($\mathbf{f}_{multi}$) are $256$, $256$ and $768$, respectively. In the cross-attention mechanism, $d_{k}$ is set to $768$ as well, therefore matching the output of the text encoder. We initialize the encoder-decoder architecture with the corresponding pre-trained modules from BLIP~\cite{li2022blip}. Since all BLIP models are publicly available, we choose the ``BLIP w/ ViT-B and CapFilt-L'' checkpoint for initialization. This model was pre-trained on 129M noisy image-text pairs using CapFilt-L, a captioning and filtering method~\footnote{\url{https://github.com/salesforce/BLIP}}. In Section~\ref{sec:backbone}, we also integrate knowledge into CLIP~\cite{radford2021learning}. For the CLIP-based model, we use ViT-B/32.

All the experiments were run on a single NVIDIA 4090. We train all the KTIR models for 10 epochs for each dataset, with a batch size of 16. We use AdamW as the optimizer and a cosine learning rate scheduler to adjust the learning rate during training. The starting learning rate is 5e-6, with a weight decay of 0.05. The temperature parameter ($\tau$) is initialized with 0.07 and is learnable during the training process. The CLIP-based models in Section~\ref{sec:backbone} are trained for 30 epochs with a batch size of 64. We set the maximum number of triplets ($m$) as 5 for all the experiments, unless stated. The loss weights $w_1$ and $w_2$ are set to 1 experimentally. Their effect will be studied in Section~\ref{sec:ablation}.

\subsection{Baseline Methods}
We compare KTIR with the following state-of-the-art methods in text-image retrieval. Based on the datasets that the model was trained on, there are two categories: supervised training methods and pretraining-finetuning methods.

\subsubsection{Supervised training methods} This category includes methods that only use the corresponding datasets to train the model. This means that the training samples are limited to the dataset. 

\begin{itemize}
    \item \textbf{AMFMN}~\cite{yuan2021exploring} employs multi-scale self-attention to extract the visual features and guide the text representation.
    \item \textbf{CMFM-Net}~\cite{yu2022text} uses a GNN to model the object and relations in the text.
    \item \textbf{GaLR}~\cite{yuan2022remote} utilizes an attention-based multi-level module to fuse global and local features extracted by a CNN and a GNN, respectively. In addition, GaLR involves a post-processing stage based on a rerank algorithm. 
    \item \textbf{KCR}~\cite{mi2022knowledge} proposes to use external knowledge triplets to expand the text content. It is the early version of KTIR. Note that, different from other methods, the text encoder is frozen in KCR.
\end{itemize}

\subsubsection{Pretraining-finetuning methods} With the development of vision-language pretraining (VLP), some methods are trained on other larger datasets and can be fine-tuned for remote sensing text-image retrieval. 

\begin{itemize}
    \item \textbf{MLT}~\cite{al2022multilanguage} a multilingual retrieval method that is based on the CLIP~\cite{radford2021learning} structure and pre-trained weights. MLT is fine-tuned on each text-image retrieval dataset separately. We report the results of the single-language mode. 
    \item \textbf{RemoteCLIP}~\cite{liu2023remoteclip} is a find-tuned CLIP model on 17 remote sensing datasets including UCM-Caption, RSICD and RSITMD and showed competitive performance on several remote sensing tasks. We report the best results of RemoteCLIP on each dataset.
    \item \textbf{GeoRSCLIP}~\cite{zhang2023rs5m} is another fine-tuned version of the CLIP model. It was pre-trained on the RS5M dataset with 5 million remote sensing images. We report the results from the best-performing model which is fine-tuned on the combination of RSICD and RSITMD datasets after trained on the the RS5M dataset.
    \item \textbf{BLIP (KTIR-base)}~\cite{li2022blip} is a VLP framework which transfers to both vision-language understanding and generation tasks. Here we fine-tune BLIP separately on the three datasets to ensure a fair comparison.
\end{itemize}

\begin{table}
\renewcommand{\arraystretch}{1.5}
\centering
\caption{Experimental results on the UCM-Caption dataset when using different knowledge sources. Methods with external knowledge are reported with gray background. The best performances are marked in \textbf{bold} and the second best are \underline{underlined}.}
\resizebox{1 \columnwidth}{!}{
\begin{tabular}{cccccccc}
\hline
\multirow{2}{*}{Knowledge Source} & \multicolumn{3}{c}{Text-Image Retrieval} & \multicolumn{3}{c}{Image-Text Retrieval} & \multirow{2}{*}{mR}\\
\cline{2-7}
 & R@1 & R@5 & R@10 & R@1 & R@5 & R@10 & \\
\hline
BLIP (KTIR-base) & \underline{19.43} & \underline{64.95} & 95.15 & 20.00 & 60.95 & {85.24} & 57.62 \\   
\rowcolor{mygray} {KTIR (GPT-4)} & {19.42} & {\textbf{66.19}} & {\textbf{95.90}} & {19.52} & {60.00} & {\underline{85.71}} & {57.79} \\
\rowcolor{mygray} KTIR (RSKG) & 18.19 & 63.62 & \textbf{95.90} & \textbf{23.33} & 60.48 & \textbf{87.14} & 58.11 \\
\rowcolor{mygray} KTIR (ConceptNet) & 18.00 & {64.86} & 94.57 & \textbf{23.33} & \underline{63.81} & 84.76 & \underline{58.22} \\
\rowcolor{mygray} KTIR (Combined) & \textbf{19.81} & 64.57 & \underline{95.33} & \underline{21.42} & \textbf{64.29} & \textbf{87.14} & \textbf{58.76} \\
\hline
\end{tabular}
}
\label{tab: knowledge}
\end{table}

\section{Results}
\label{sec:result}

\subsection{The Effectiveness of External Knowledge}

In our first experiment, we investigate the effectiveness of the two different knowledge sources (RSKG and ConceptNet) and of their combination. To further explore the effectiveness of different knowledge sources, we also augment KTIR using a large language model: we prompt GPT-4~\cite{achiam2023gpt} to generate relevant sentences to expand the scope of the given text, and then use these generated sentences as knowledge sentences for KTIR. To ensure a fair comparison, the number of generated sentences is kept the same as the number of knowledge sentences constructed from the other knowledge graphs, \textit{i.e.}, $m$.
In Table~\ref{tab: knowledge}, we compare KTIR with different knowledge sources with the fine-tuned BLIP modal on the UCM-Caption dataset. The results suggest external concepts can supply text information to fill the information gap between text descriptions and image content. For example, KTIR with GPT-4, RSKG and ConceptNet improve the mR of the model without knowledge by 0.17\%, 0.49\% and 0.60\%, respectively. By combining RSKG and ConceptNet, KTIR improves the performance of the baseline model even further, by 1.14\%. The GPT-4-augmented KTIR underperforms KTIR with knowledge graphs. This is likely due to generative models hallucinating irrelevant sentences about the content, along with their lack of stability and reproducibility during the generation process, which might pose challenges to the retrieval task. Comparing the three different knowledge graphs as knowledge sources, KTIR with the combination of ConceptNet and RSKG outperforms the other two sources in isolation, suggesting the effectiveness of balancing external concepts and domain-specific knowledge.
According to the experimental results, KTIR is not limited to a specific knowledge source and exhibits scalability for larger knowledge graphs. However, there is a trade-off between the size of the knowledge sources and the relevance of external knowledge. Especially, in the case of current remote sensing captioning datasets, the captions are generally short and span a limited set of concepts, which would advocate for smaller, more targeted knowledge graphs.

\begin{figure}[!t]
    \centering
    \includegraphics[width=0.9\linewidth]{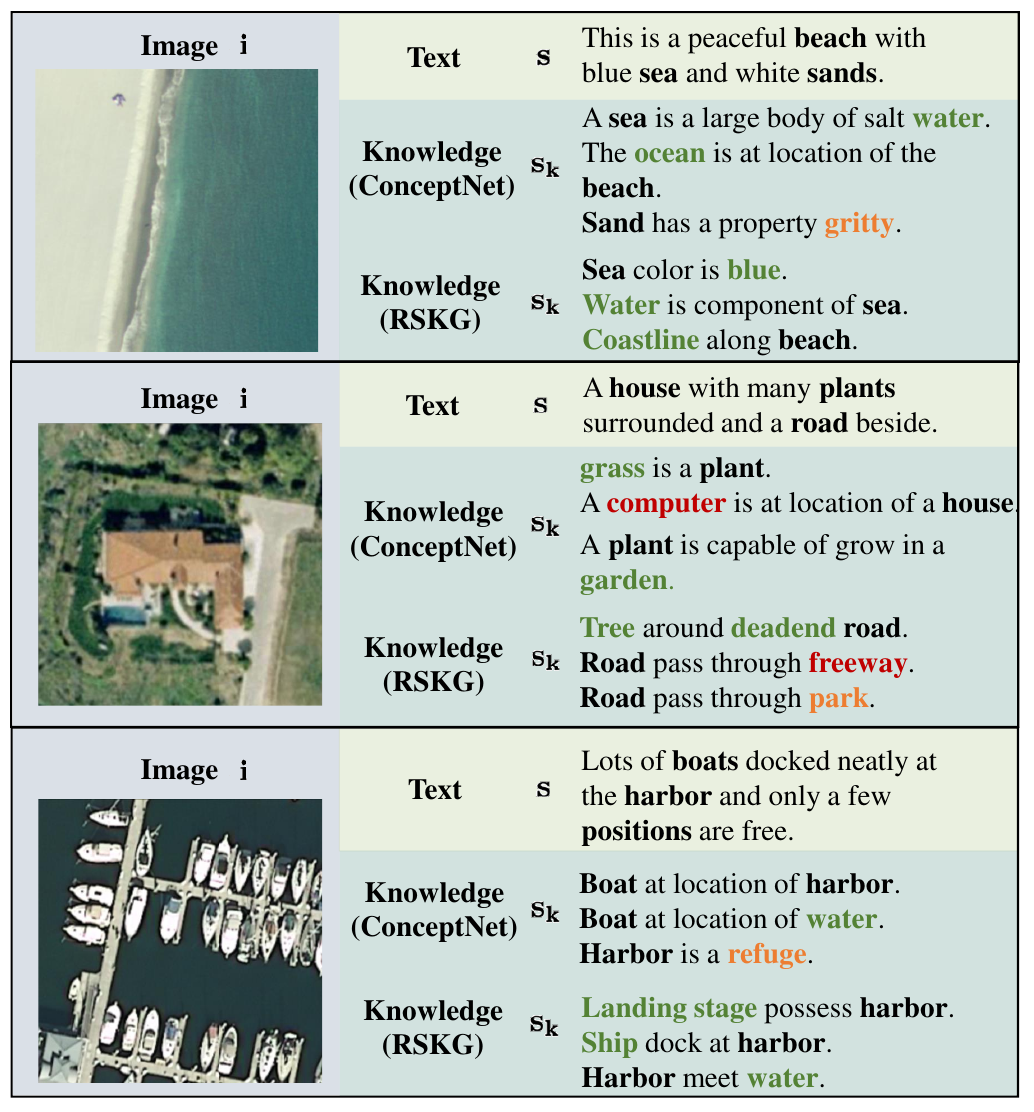}
    \caption{Examples retrieved knowledge sentences from different knowledge sources. Keywords (nouns) in the text and knowledge sentences are in \textbf{bold}. External concepts, whether highly related, somewhat related, or unrelated to the image content, are respectively marked in \textcolor{green2}{\textbf{green}}, \textcolor{orange}{\textbf{orange}}, and \textcolor{red}{\textbf{red}}. If there are more than 3 knowledge sentences, we randomly select 3 sentences from the knowledge extraction results. The colors are manually annotated.}
    \label{fig:knowledge}
\end{figure}

\label{sec:knowledge}
\paragraph{\textbf{Comparing the knowledge sentences by source}}
A few examples of the retrieved knowledge sentences are shown in Fig.~\ref{fig:knowledge}. Based on the keywords (in bold in the text), the knowledge retrieval process introduces relevant concepts (in green) to expand the text scope. From the figure we can see that both sources bring related objects or relations into knowledge sentences $\mathbf{s_{k}}$. For example, for the set of keywords \textit{(beach, sea, sands)} in the first image, both sources add \textit{water} to the knowledge sentences. Compared between those two sources, the knowledge sentences from the hand-crafted knowledge graph are more specific to the remote sensing field, while the information from the ConceptNet is more generic but diverse. For instance, in the second example, ConceptNet adds \textit{grass} and \textit{garden} that are highly related to the scene but it also includes \textit{computer} that cannot be seen from the image. When combining two knowledge sources, the corresponding triplet can be filtered out as \textit{computer} and \textit{house} are not in RSKG.

\begin{figure}[!t]
    \centering
    \includegraphics[width=0.9\linewidth]{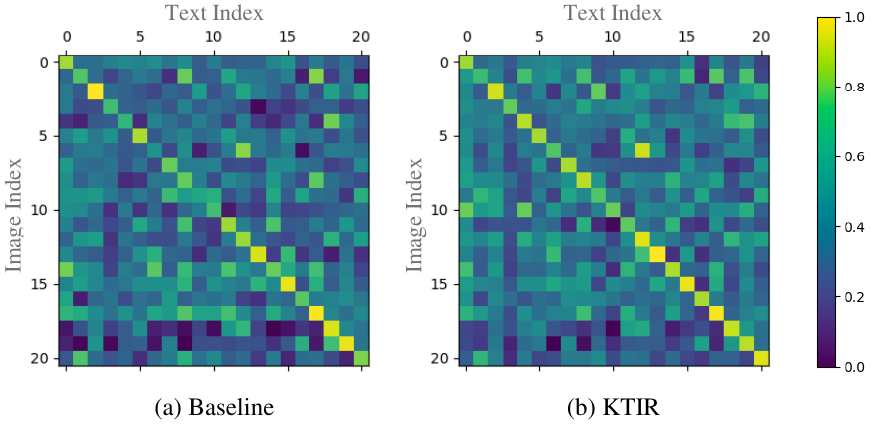}
    \caption{The cosine similarity scores for image-text retrieval of 21 image and text pairs from the UCM-Caption dataset sampled from different scene categories. The horizontal axis represents the text index, and the vertical axis represents the image index.}
    \label{fig:similarity}
\end{figure}

\begin{figure}[!t]
    \centering
    \includegraphics[width=0.98\linewidth]{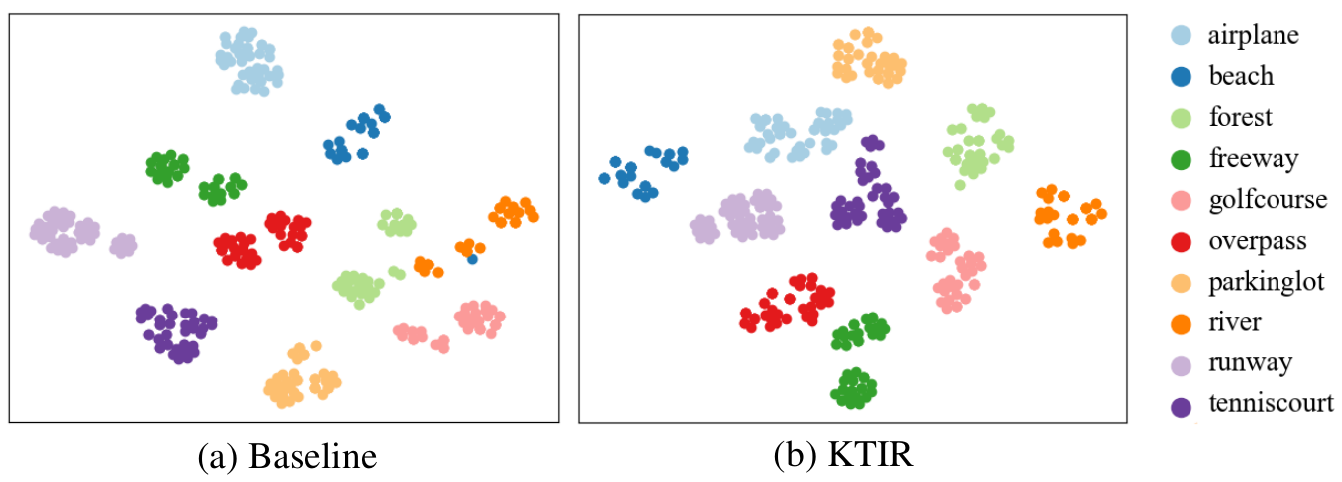}
    \caption{The t-SNE visualization of the text features of the baseline and of the proposed KTIR method on the UCM-Caption dataset. For visualization, we randomly select 10 scene categories out of 21 categories in total.}
    \label{fig:embedding}
\end{figure}

From now on, we only consider combining RSKG and ConceptNet as the knowledge source, so `KTIR' in the following sections denotes KTIR (Combined). Unless specified, otherwise, `Baseline' denotes BLIP (KTIR-base), which is fine-tuned on the specific dataset without external knowledge.

\begin{table*}
\tiny
\renewcommand{\arraystretch}{1.15}
\centering
\caption{Results on the UCM-Caption dataset. Methods in the first block are the supervised learning methods, while in the second block, we present the pretraining-finetuning methods. $*$ module weights frozen. Methods with external knowledge are reported with gray background. Note that we report the best results of comparing methods from the corresponding original papers on each dataset.}
\resizebox{1.8 \columnwidth}{!}{
\begin{tabular}{cccccccccc}
\hline
\multirow{2}{*}{Model} & \multirow{2}{*}{Image Encoder} & \multirow{2}{*}{Text Encoder} & \multicolumn{3}{c}{Text-Image Retrieval} & \multicolumn{3}{c}{Image-Text Retrieval} & \multirow{2}{*}{mR}\\
\cline{4-9}
& & &  R@1 & R@5 & R@10 & R@1 & R@5 & R@10 & \\
\hline 
AMFMN~\cite{yuan2021exploring} & ResNet-50 & Bi-GRU & 12.86 & 53.24 & 79.43 & 16.67 & 45.71 & 68.57 & 46.08 \\ 
\rowcolor{mygray} KCR~\cite{mi2022knowledge} & ResNet-18 & Transformer$^{*}$ & 17.24 & 56.95 & 81.14 & 11.90 & 48.57 & 71.43 & 47.87 \\ 
\hline
MLT~\cite{al2022multilanguage} & ViT-B-32 & Transformer & 19.33 & 64.00 & 91.42 & 19.04 & 53.33 & 77.61 & 54.12 \\
RemoteCLIP~\cite{liu2023remoteclip} & ViT-B-32 & Transformer & 18.67 & 61.52 & 94.29 & 20.48 & 59.85 & 83.33 & 56.36 \\
BLIP (KTIR-base)~\cite{li2022blip} & ViT-B-16 & Transformer & 19.43 & \textbf{64.95} & 95.15 & 20.00 & 60.95 & 85.24 & 57.62 \\
\rowcolor{mygray} KTIR (Combined) & ViT-B-16 & Transformer & \textbf{19.81} & 64.57 & \textbf{95.33} & \textbf{21.42} & \textbf{64.29} & \textbf{87.14} & \textbf{58.76} \\
\hline
\end{tabular}
}
\label{tab:ucm}
\end{table*}

\begin{table*}[t]
\tiny
\renewcommand{\arraystretch}{1.15}
\centering
\caption{Rresults on the RSICD dataset. (See the caption of Table~\ref{tab:ucm} for details)}
\resizebox{1.8 \columnwidth}{!}{
\begin{tabular}{cccccccccc}
\hline
\multirow{2}{*}{Model} & \multirow{2}{*}{Image Encoder} & \multirow{2}{*}{Text Encoder} & \multicolumn{3}{c}{Text-Image Retrieval} & \multicolumn{3}{c}{Image-Text Retrieval} & \multirow{2}{*}{mR}\\
\cline{4-9}
& & & R@1 & R@5 & R@10 & R@1 & R@5 & R@10 & \\
\hline 
AMFMN~\cite{yuan2021exploring} & ResNet-50 & Bi-GRU & 4.90 & 18.28 & 31.44 & 5.39 & 15.08 & 23.40 & 16.42 \\ 
CMFM-Net~\cite{yu2022text} & ResNet-18 & Bi-GRU & 5.31 & 18.57 & 30.03 & 5.40 & 18.66 & 28.55 & 17.75 \\
GaLR~\cite{yuan2022remote} & ResNet-18 & Bi-GRU & 4.69 & 19.48 & 32.13 & 6.59 & 19.85 & 31.04 & 18.96  \\ 

\rowcolor{mygray} KCR~\cite{mi2022knowledge} & ResNet-18 & Transformer$^{*}$ & 5.84 & 22.31 & 36.12 & 4.76 & 18.59 & 27.20 & 19.14 \\ 
\hline
MLT~\cite{al2022multilanguage} & ViT-B-32 & Transformer & 9.14 & 28.96 & 44.59 & 10.70 & 29.64 & 41.53 & 27.42 \\
RemoteCLIP~\cite{liu2023remoteclip} & ViT-L-14 & Transformer & 14.73 & 39.93 & 56.58 & 18.39 & 37.42 & 51.05 & 36.35 \\
GeoRSCLIP~\cite{zhang2023rs5m} & ViT-B-32 & Transformer & 15.59 & 41.19 & 57.99 & 21.13 & 41.72 & 55.63 & 38.87 \\
BLIP (KTIR-base)~\cite{li2022blip} & ViT-B-16 & Transformer & \textbf{20.64} & 47.50 & 63.13 & 25.89 & 47.48 & 58.92 & 43.93 \\ 
\rowcolor{mygray} KTIR (Combined) & ViT-B-16 & Transformer & 20.55 & \textbf{48.67} & \textbf{63.70} & \textbf{26.08} & \textbf{49.77} & \textbf{62.49} & \textbf{45.21} \\
\hline
\end{tabular}}
\label{tab:rsicd}
\end{table*}

\paragraph{\textbf{The impact of external knowledge}}
To further investigate the impact of external knowledge, we show the cosine similarity scores between image and text feature in Fig.~\ref{fig:similarity}.
We randomly select one image-text pair per scene category from 21 categories in the UCM-Caption dataset. For both results from the baseline model and KTIR, there is a clear diagonal pattern, which means in general the paired image and text have higher similarity scores. The paired image and text have relatively higher scores compared with the unpaired ones in the results from KTIR, which indicates that the embeddings become more discriminative by adding external knowledge.

The t-SNE~\cite{van2008visualizing} visualization of text features from the baseline model and the proposed KTIR are shown in Fig.~\ref{fig:embedding}. We randomly selected 10 scene categories and visualized all the text embeddings from the text set of the UCM-Caption dataset. After adding external knowledge, we observed three characteristics: 1) Within categories, the features become more compact after adding knowledge. That is probably because the knowledge added to similar scenes is similar (See the orange example of \textit{river}). 2) There are more sub-centers in each category, which indicates by adding knowledge the model also learns to identify small differences within categories (See the purple example of \textit{tenniscourt}). 3) Semantically similar categories get closer in the embedding space. For example, \textit{runway} and \textit{airplane}, \textit{golfcourse} and \textit{forest}, \textit{overpass} and \textit{runway}, \textit{etc}. This indicates that by integrating external knowledge, the model learns a more representative semantic space for the remote sensing domain.

\begin{table*}[t]
\tiny
\renewcommand{\arraystretch}{1.16}
\centering
\caption{Results on the RSITMD dataset. (See the caption of Table~\ref{tab:ucm} for details)}
\resizebox{1.8 \columnwidth}{!}{
\begin{tabular}{cccccccccc}
\hline
\multirow{2}{*}{Model} & \multirow{2}{*}{Image Encoder} & \multirow{2}{*}{Text Encoder} & \multicolumn{3}{c}{Text-Image Retrieval} & \multicolumn{3}{c}{Image-Text Retrieval} & \multirow{2}{*}{mR}\\
\cline{4-9}
&  & & R@1 & R@5 & R@10 & R@1 & R@5 & R@10 & \\
\hline 
CMFM-Net~\cite{yu2022text} & ResNet-18 & Bi-GRU & 10.00 & 32.83 & 47.21 & 10.84 & 28.76 & 40.04 & 28.28\\
AMFMN~\cite{yuan2021exploring} & ResNet-50 & Bi-GRU &  11.51 & 34.69 & 54.87 & 10.63 & 24.78 & 41.81 & 29.72 \\ 
GaLR~\cite{yuan2022remote} & ResNet-18 & Bi-GRU & 11.15 & 36.68 & 51.68 & 14.82 & 31.64 & 42.48 & 31.41  \\ 
\hline
MLT~\cite{al2022multilanguage} & ViT-B-32 & Transformer & 17.61 & 49.73 & 66.59 & 19.69 & 40.26 & 54.42 & 41.38 \\
RemoteCLIP~\cite{liu2023remoteclip} & ViT-L-14 & Transformer & 23.76 & 59.51 & 74.73 & 28.76 & 52.43 & 63.94 & 50.52 \\
GeoRSCLIP~\cite{zhang2023rs5m} & ViT-B-32 & Transformer & 25.04 & 57.88 & 74.38 & 32.30 & 53.32 & \textbf{67.92} & 51.81\\
BLIP (KTIR-base)~\cite{li2022blip} & ViT-B-16 & Transformer & 30.66 & 60.62 & 76.19 & \textbf{34.51} & 52.88 & 65.49 & 53.39\\
\rowcolor{mygray} KTIR (Combined) & ViT-B-16 & Transformer & \textbf{31.46} & \textbf{62.92} & \textbf{76.59} & 34.29 & \textbf{55.31} & 65.04 & \textbf{54.27}\\
\hline

\end{tabular}}
\label{tab:rsitmd}
\end{table*}

\subsection{Results on Remote Sensing Benchmarks}

\paragraph{UCM-Caption (Table \ref{tab:ucm})}
Compared with supervised learning methods, pretraining-finetuning methods have better performance, which indicates the effectiveness of the latter learning strategy. KTIR gains 10.89\% on mR over the knowledge-aware baseline KCR confirming the superior representation ability brought from the pretraining stage.
mR shows a 1.14\% increase due to the introduction of relevant external information from knowledge sources compared to BLIP. The performance improvements indicate that adding external knowledge also benefits bridging the gap between natural images and remote sensing images.

\paragraph{RSICD (Table \ref{tab:rsicd})}
KTIR achieves the best performance over all the competitors. With equivalent or relatively smaller backbones, KTIR outperforms the state-of-the-art methods (6.34\% on mR with respect to the best-performing methods). 
Note that compared to GeoRSCLIP, which was pre-trained on 5M remote sensing images and fine-tuned on the RSICD dataset, the proposed KTIR is only fine-tuned on the RSICD dataset. Compared with the BLIP (KTIR-base) model, adding knowledge improves the results by 1.28\% on mR, which also indicates the effectiveness of our approach: the knowledge related to specific image content may help bridge the gap between pretraining data and fine-tuning data.

\begin{figure*}[!t]
    \centering
    \includegraphics[width=0.85\linewidth]{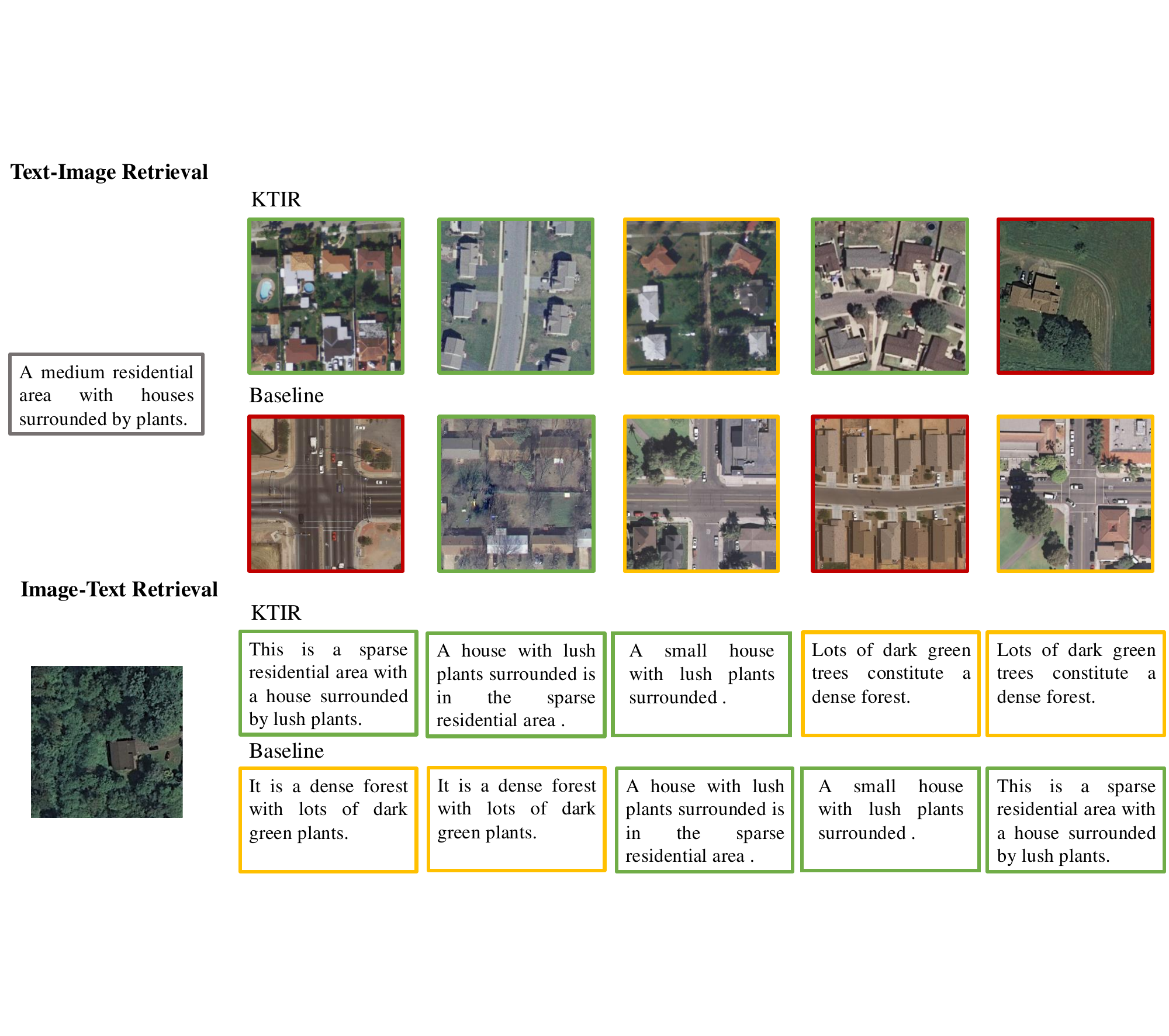}
    \caption{Examples of KTIR's and the baseline model (BLIP)'s top 5 retrieval results on the UCM-Caption dataset for both text-image retrieval and image-text retrieval. Texts or images in \textcolor{green2}{\textbf{green}} boxes are predictions that correspond to the ground truth. If the Texts or images are in \textcolor{yellow1}{\textbf{yellow}} boxes, these results are not annotated as the ground truth but are regarded as reasonable predictions according to human evaluation. Texts or images in \textcolor{red}{\textbf{red}} are semantically incorrect retrieval results. Note that retrieved images or texts matching all the semantic concepts are considered reasonable results, while results missing one or more semantic concepts are considered incorrect results.}
    \label{fig:examples}
\end{figure*}

\begin{figure}[!t]
    \centering
    \includegraphics[width=0.88\linewidth]{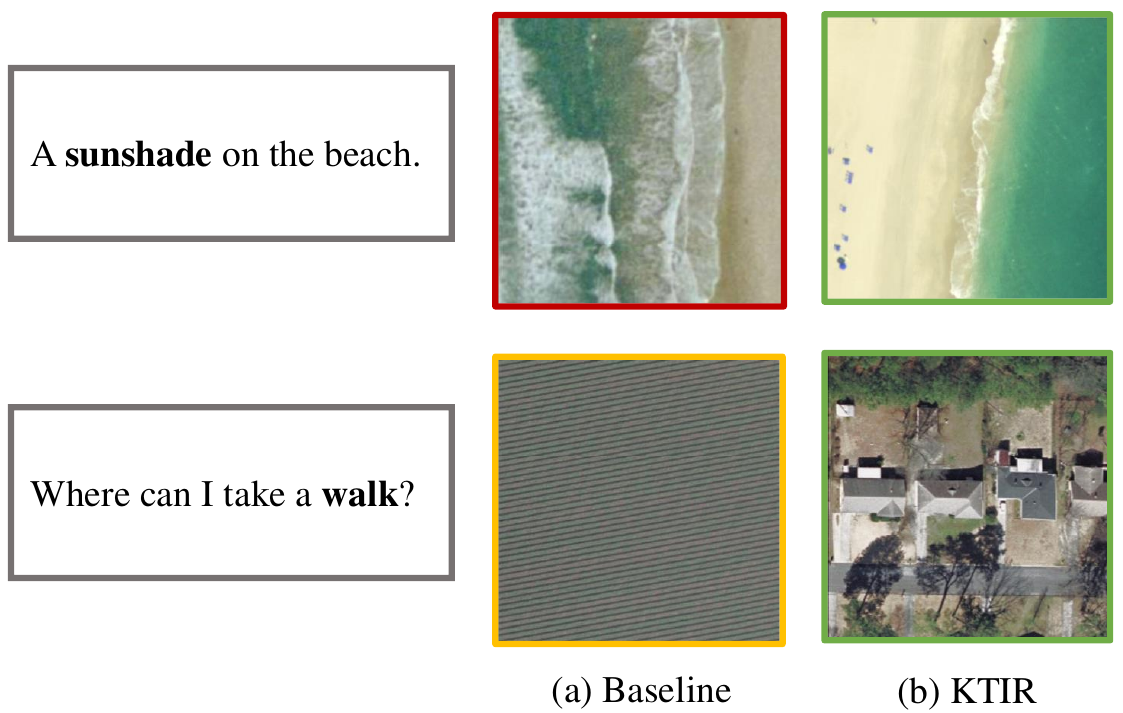}
    \caption{Open-set text-image retrieval on the UCM-Caption dataset. For each example, the text query is on the left and the Top-1 image retrieval results of the baseline model and KTIR are shown on the right. Unseen concepts during training are marked in \textbf{bold}.}
    \label{fig:zero}
\end{figure}

\paragraph{RSITMD (Table \ref{tab:rsitmd})}
Images in the RSITMD dataset have more diverse and fine-grained captions than those in the RSICD dataset. KTIR shows competitive performance on both text-image retrieval and image-text retrieval, leading to an overall improvement on mR of 2.46\% with respect to GeoRSCLIP, and larger gaps with respect to all the other baselines. In terms of text-image retrieval, KTIR outperforms all the other comparison methods. For image-text retrieval, GeoRSCLIP has a better performance than KTIR in terms of R@10. The potential explanation is the insufficient training of the image encoder as we extract pre-trained weights from natural images and fine-tune the model directly on a specific remote sensing dataset. The performance gain of KTIR over the baseline model (0.88\%) is also smaller compared to the performance gain on the RSICD dataset (1.28\%), which may be due to the fact that the descriptions in the RSITMD dataset are more diverse than in the RSICD dataset, and therefore there may be less benefit from adding external knowledge.

\subsection{Qualitative Analysis}

Some visual examples of retrieval are shown in Fig.~\ref{fig:examples}: KTIR not only has higher retrieval performance but also orders the candidate captions or images better in the ranking. 
In the text-image examples, the images retrieved are more reasonable. Similarly, in the image-text example, integrating external information redirects the model's attention towards a more prominent part of the image.

We also perform an open-set text-image retrieval experiment, where some concepts in the caption are not seen by the model during training. Results are shown in Fig.~\ref{fig:zero}. By integrating external knowledge, the model can generalize well to unseen captions. When encountering unseen concepts (\textit{e.g.,} \textit{sunshade} and \textit{walk}), KTIR links them with known objects based on commonsense knowledge while the baseline fails to retrieve reasonable results.

\subsection{Ablations and Parameter Analysis}
\label{sec:ablation}
We design ablation studies to provide further insights into the proposed KTIR method, including the analysis of the model architecture and the loss functions. The results are shown in Table~\ref{tab:ablation}.
Starting from the image encoder ($\operatorname{ViT}$) and text-only mode of text encoder ($\operatorname{BERT}_{text}$) with knowledge-aware contrastive loss ($\mathcal{L}_{con}$) (row 1), we add the multimodal mode of the text encoder ($\operatorname{BERT}_{multi}$) and the knowledge-aware matching loss ($\mathcal{L}_{mat}$) (row 2). Then we add the cross-attention layer to fuse the captions and knowledge (row 3). The impact of the cross-attention layer can be seen in row 2 and row 3. By using a cross-attention mechanism to fuse the knowledge embedding and text embedding, the model slightly improves its performance (0.55\% on mR). The model gains 1.05\% by using two losses with respect to the contrastive objective only indicating knowledge emphasizes the effectiveness of the text-image matching objective proposed in BLIP. This may be due to the lack of scene and description diversity in the specific dataset. In this case, adding additional pairing information helps model training.

\begin{table}[t]
\renewcommand{\arraystretch}{1.2}
\centering
\caption{Ablation studies of model architecture and loss functions on the UCM-Caption dataset. CrossAtt denotes the cross-attention layer to fuse the knowledge embedding and the caption embedding. $\mathrm{mR}_{t2i}$ and $\mathrm{mR}_{i2t}$ denote the average of R@$k$ in text-image and image-text retrieval, respectively.}
\resizebox{0.8 \columnwidth}{!}{
\begin{tabular}{cccccc}
\hline
CrossAtt & $\mathcal{L}_{\mathrm{mat}}$ & $\mathcal{L}_{\mathrm{con}}$ & mR$_{t2i}$ & mR$_{i2t}$ & mR\\
\hline 
 &  & \checkmark & 58.29 & 56.03 & 57.16 \\
 & \checkmark & \checkmark & \textbf{60.54} & 55.87 & 58.21  \\
 \checkmark & \checkmark & \checkmark &  59.90 & \textbf{57.62} & \textbf{58.76} \\
\hline
\end{tabular}
}
\label{tab:ablation}
\end{table}

\begin{table}[t]
\renewcommand{\arraystretch}{1.2}
\centering
\caption{Parameter analysis of different loss weight settings on UCM-Caption dataset. ``$w_1$'' and ``$w_2$'' denote the weights of two loss terms in Eq.~\ref{eq:loss}.}
\resizebox{0.8 \columnwidth}{!}{
\begin{tabular}{ccccc}
    \hline
    {$w_1$ ($\mathcal{L}_{con}$)} & {$w_2$ ($\mathcal{L}_{mat}$)} & {mR$_{t2i}$} & {mR$_{i2t}$} & {mR}\\
    \hline 
    {$0.5$} & {$1$} & {\textbf{59.97}} & {56.67} & {58.32} \\
    {$1$} & {$0.5$} & {59.24} & {\textbf{58.25}} & {58.75} \\
    {$1$} & {$1$} & {59.90} & {57.62} & {\textbf{58.76}} \\ 
    \hline
    \end{tabular}
}
\label{tab:weight}
\end{table}

\begin{table}[t]
\renewcommand{\arraystretch}{1.2}
\centering
\caption{Parameter analysis of the number of triplets on UCM-Caption dataset. ``$m$'' demotes the number of triplets per caption.}
\resizebox{0.57 \columnwidth}{!}{
\begin{tabular}{cccc}
\hline
$m$ & mR$_{t2i}$ & mR$_{i2t}$ & mR\\
\hline 
$m=1$ &  \textbf{60.16} & 55.40 & 57.78 \\
$m=3$ &  59.68 & 56.66 & 58.17 \\
$m=5$ &  59.90 & \textbf{57.62} & \textbf{58.76} \\ 
{$m=7$} & {59.46} & {57.46} & {58.46} \\
{$m=10$} & {59.58} & {55.87} & {57.73} \\ 

\hline
\end{tabular}
}
\label{tab:number}
\end{table}

To study the impact of the loss weights (Eq.~\ref{eq:loss}), we tested the model with different values of $(w_1$, $w_2)$ : $(0.5, 1)$, $(1, 0.5)$ and $(1, 1)$. The results in Table~\ref{tab:weight} suggest that the best performance is achieved when the two losses have the same weights ($w_1=w_2=1$). When $w_1>w_2$, the importance of the contrastive loss component increases, leading to a 0.63\% improvement in image-text retrieval performance. This indicates the impact of the contrastive objective in retrieving the right text based on the image content. When $w_2>w_1$, the performance of text-image retrieval improves instead (mR$_{t2i}$ increases 0.07\%). This suggests that the matching objective plays a significant role in retrieving images based on text descriptions. These findings also align with the trends observed in the loss ablation results (see rows $1$ and $2$ in Table~\ref{tab:ablation}).

The number of triplets ($m$) that are used during retrieval is an important parameter of KTIR: it controls the proportion of information extracted from external knowledge sentences and fed into the final text. We varied $m$ to 1, 3, 5, 7 and 10 to study the impact of this parameter on the UCM-Caption dataset in Table~\ref{tab:number}. We randomly select $m$ triplets from the triplet lists to generate the knowledge sentences if the total number of triplets is more than $m$, otherwise, we keep the original number of the triplets. The results show that the balance between original information and external knowledge is achieved when $m=5$. The mR improves 2.22\% compared to $m=1$ on the image-text retrieval ($i2t$) and 0.98\% overall. The opposite behaviour is observed in the text-image retrieval ($t2i$). As the number of triplets rises from 1 to 5, the effectiveness of text-image retrieval decreases. This suggests that as text representations become more varied, the task of retrieving the same image from various texts becomes more challenging, whereas discerning text based on images becomes simpler. When $m>5$, the performance of image-text retrieval drops more dramatically than in the case of text-image retrieval. This is likely due to the fact that, as the number of knowledge sentences increases, the information from the caption contributes less to the final text representation and distinguishing text descriptions becomes more challenging.

\subsection{Triplet selection methods}\label{ssec:triplet_selection}

\begin{table}[t]
\renewcommand{\arraystretch}{1.2}
\centering
\caption{Experimental results on the UCM-Caption dataset with different triplet selection methods.}
\resizebox{0.8 \columnwidth}{!}{
\begin{tabular}{cccc}
\hline
{$m$} & {mR$_{t2i}$} & {mR$_{i2t}$} & {mR}\\
\hline 
{Relevance with the captions} & {59.05}  & {57.30} & {58.18} \\
{Diversity among triplets} & {59.56} & {56.51} & {58.04} \\
{Random selection} &  {\textbf{59.90}} & {\textbf{57.62}} & {\textbf{58.76}} \\ 

\hline
\end{tabular}
}
\label{tab:select}
\end{table}

Selecting knowledge triplets at random could be sub-optimal. To understand the impact of different triplet selection methods, we tested KTIR with three approaches:
\begin{itemize}
    \item \textbf{Relevance to the caption}. We first measure the semantic similarity between the triplets and their corresponding caption using a pre-trained sentence BERT model~\cite{reimers2019sentence}. To maximize the use of relevant triplets to expand the caption's content, we then rank the triplets in reverse order based on the similarity scores and select the top $m$ ones, \textit{i.e.}, those with the largest semantic distance from the caption.
    \item \textbf{Diversity among triplets}. We randomly select the first triplet, and then measure the semantic relevance score between the remaining triplets and the selected one. To maximize diversity among the selected triplets, we then rank the remaining ones in reverse order based on the semantic relevance scores and choose the top $m$ ones.
    \item \textbf{Random selection}. We shuffle the retrieved triplet list and randomly select $m$ triplets from all the possible ones.
\end{itemize}
We use the above triplet selection methods to select triplets and train the model separately on the UCM-Caption dataset. The experimental results are reported in Table~\ref{tab:select}. Although relevance is the basis of the knowledge-aware method to retrieve and extract information from knowledge sources, we observe that an active selection of the knowledge triplet does not seem to improve the results numerically: random selection yields the best overall performance (gains of on average 0.65\% in terms of mR). On one hand, the success of random selection may be attributed to the fact that the exploration of the knowledge graph is limited since knowledge retrieval is restricted to the one-step neighborhood on the graph. Therefore, even with random selection, the knowledge triplets maintain high semantic relevance to the captions. On the other hand, random selection might introduce more triplet combinations and higher error tolerance to the retrieval process which could also benefit the overall effectiveness. The results also encourage future research on strategies to explore knowledge sources that balance semantic relevance and information diversity.

\subsection{Backbone Analysis}
\begin{figure}[!t]
    \centering
    \includegraphics[width=0.9\linewidth]{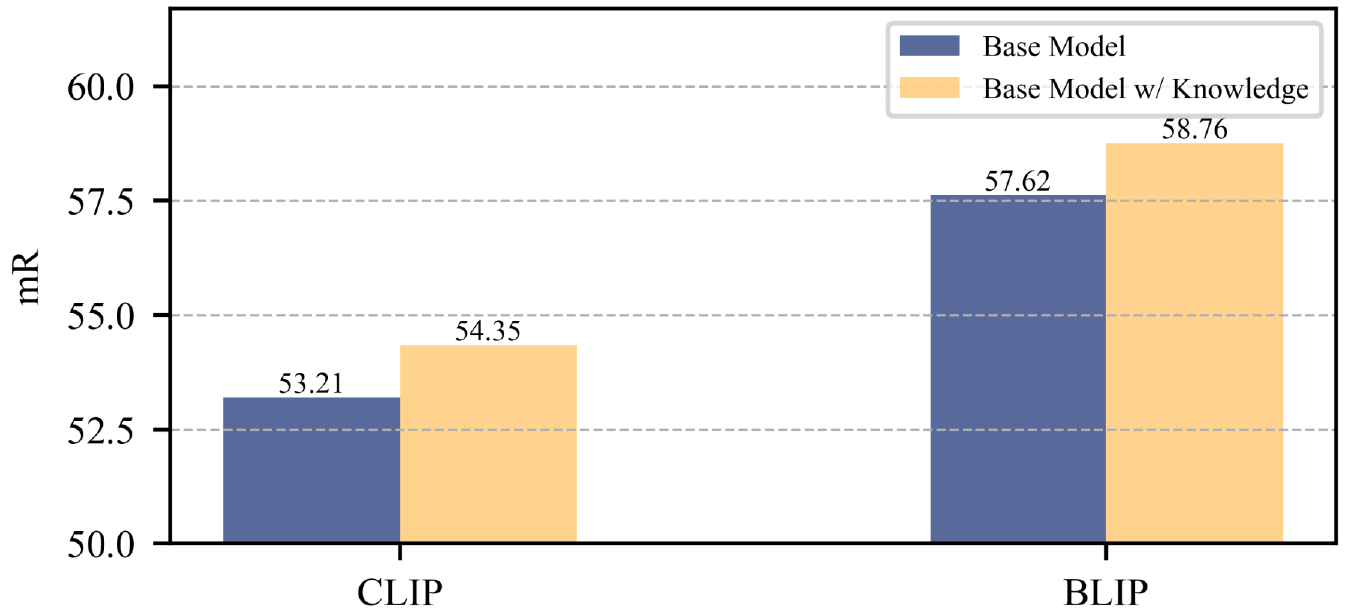}
    \caption{Comparasion of using CLIP~\cite{radford2021learning} and BLIP~\cite{li2022blip} the base models on UCM-Caption dataset. mR is reported in the figure.}
    \label{fig:backbone}
\end{figure}

To further demonstrate the effectiveness of external knowledge, we integrate knowledge into different backbone models including CLIP~\cite{radford2021learning} and BLIP~\cite{li2022blip} on the UCM-Caption dataset. For the CLIP-based model, we only do fast knowledge integration (combining the captions with corresponding knowledge sentences as text inputs), since there is no cross-attention module in the text encoder. Note that because of these architecture differences, the knowledge-aware text-image matching loss ($\mathcal{L}_{mat}$) cannot be used in CLIP-based model. The results on UCM-Caption in Fig.~\ref{fig:backbone} show that adding external knowledge leads to better performance for both backbones, which further indicates the generalizability of knowledge-aware models. Moreover, the results also suggest that domain-specific knowledge facilitates the adaptation of general pre-trained models to the remote sensing domain.

\label{sec:backbone}

\section{Conclusion}
\label{sec:conclusion}
Retrieving remote sensing images from text queries is appealing but complex since retrieval needs to be both visual and semantic. To address the information asymmetry between images and texts, we therefore propose a Knowledge-aware Text-Image Retrieval (KTIR) method. By integrating relevant information from external knowledge sources, the model enriches the text scope and alleviates the potential ambiguity to better match texts and images. Extensive experiments on three datasets show that KTIR outperforms all competitors and creates representative semantic space for remote sensing images. Supportive analysis further demonstrates the effectiveness and potential generalization capabilities of the knowledge-aware methods to unseen concepts and various backbones. We hope these results will encourage future research beyond the world of pixels and embrace new sources of knowledge towards better image retrieval systems.

\ifCLASSOPTIONcaptionsoff
  \newpage
\fi

{\small
\bibliographystyle{IEEEtran}
\bibliography{reference}
}

\clearpage
\clearpage

\end{document}